\documentclass{evolang12}
\usepackage{graphicx}
\usepackage{url}
\usepackage{authblk}
\usepackage{amsthm}

\begin{document}

\title{Distribution-based Prediction of the Degree of Grammaticalization for German Prepositions}
\author[*]{DOMINIK SCHLECHTWEG}
\author[ ]{SABINE SCHULTE IM WALDE}
\affil[*]{dominik.schlechtweg@gmx.de}
\affil[ ]{Institute for Natural Language Processing, University of Stuttgart}

\maketitle

\section{Introduction}

Grammaticalization refers to the diachronic ``development from lexical to grammatical forms, and from grammatical to even more grammatical forms'' \shortcite[p.~32]{heineKut:2007}. It is assumed to go along with \textit{desemanticization}, a process of losing descriptive meaning \shortcite{Heine:2003,Bybee:2015aa,lehmann2015thoughts,heineKut:2007}.
For instance, the German noun \textit{Trotz} `defiance' acquired a less specific, grammatical meaning as preposition, cf. \textit{trotz des Sturms} `despite the storm', in which the original descriptive meaning is lost. The resulting prepositional meaning is semantically more \underline{general} and often highly \underline{polysemous} \shortcite[cf.~p.~134,~151]{DiMeola00}. In addition, grammatical categories show a high degree of \underline{obligatoriness} \shortcite[cf.~p.~39f.]{DiMeola00}, i.e., they must be specified in a sentence \shortcite[cf.~p.~14]{lehmann2015thoughts}. These three highlighted properties of grammaticalized expressions (generality, polysemy, obligatoriness) have a directly observable impact on their contextual distributions: they are used in a greater number of contexts \shortcite{Weeds:2003,heineKut:2007,Santus:2014,Bybee:2015aa,schlechtweg-EtAl:2017:CoNLL}. Together with the assumption that grammaticalization is a continuous process \shortcite[cf.~p.~68]{DiMeola00}, these observations motivate our central hypothesis: 

\begin{description}
\item[Hypothesis] The degree of grammaticalization of an expression correlates with the unpredictability of its context words (contextual dispersion). 
\end{description}

\section{Method}

In computational linguistics, a prominent corpus-based measure of contextual dispersion is word entropy \shortcite{Hoffman:2013,Santus:2014}. We exploit this measure in order to test our central hypothesis. First, we create a test set of German prepositions with different degrees of grammaticalization; we then (i) compute Spearman's rank-order correlation coefficient ($\rho$) between test set and word entropy scores, and (ii) use Average Precision (AP) to measure how well the scores distinguish between degrees.

For computing word entropy we induce a Distributional Semantic Model with window size 2 from a part of the SdeWaC corpus \shortcite{faasseckart2012} with approx. 230 million tokens. Low-frequency and functional words are deleted, and every token is replaced by its lemma plus POS-tag. For comparison, we also compute other quantitative measures of different aspects of contextual dispersion: word frequency and the number of context types.\footnote{Code: \url{https://github.com/Garrafao/MetaphoricChange}.}

\section{Test Set}

\shortciteA{DiMeola00} distinguishes between (i) prepositions with the form of a content word (e.g., \textit{trotz}), (ii) prepositions with the form of a syntactic structure (e.g., \textit{am Rande}) and (iii) prepositions with the form of a function word (e.g., \textit{vor}). Prepositions in (i) and (ii) show a low to medium degree of grammaticalization, while the ones in (iii) show a high degree (cf.~p.~60). We focus on prepositions with the form of a PP from (ii), because \citeauthor{DiMeola00} provides fine-grained distinctions, and (iii), to exploit a wide range of degrees. The final test set contains 206 prepositions with four degrees of grammaticalization (1: low -- 4: high).\footnote{The test set is provided together with the predicted measure scores at \url{http://evolang.org/torun/proceedings/papertemplate.html?p=169}.}

\section{Results}

Table 1 shows that there is indeed a moderate correlation between entropy and the degree of grammaticalization, but frequency and the number of context types outperform entropy (all $\rho$ significantly different from 0, $p < .01$, two-tailed, t-test). Frequency has the highest overall correlation with grammaticalization: it is only .01 different to the correlation of context types ($p = .6$, two-tailed, \citeauthor{Steiger1980}'s Z-test), but with .04 clearly different from entropy ($p = .06$). Frequency also distinguishes best between most of the degree levels; context types are generally comparable and in one case even the best predictor. Overall, the table clearly demonstrates that the more different the degrees of grammaticalization are, the better they are distinguished by the three measures.

\begin{table}[ht]
 \tablecaption{Results for predicting degrees of grammaticalization.}
{\footnotesize
\begin{tabular}{@{}ccccc@{}}
\hline
 & entropy & frequency & types \\
\hline
AP (degrees 1 vs. 2) & 0.54 & \textbf{0.56} & 0.55 \\
AP (degrees 1 vs. 3) & 0.67 & \textbf{0.68} & \textbf{0.68} \\
AP (degrees 1 vs. 4) & 0.89 & \textbf{0.92} & \textbf{0.92} \\
AP (degrees 2 vs. 3) & 0.67 & \textbf{0.69} & 0.68 \\
AP (degrees 2 vs. 4) & 0.89 & \textbf{0.92} & \textbf{0.92} \\
AP (degrees 3 vs. 4) & 0.84 & 0.87 & \textbf{0.88} \\\hline
Spearman's $\rho$ (rank) & 0.42 & \textbf{0.46} & 0.45 \\

\hline
\end{tabular}\label{tab:results}}
\end{table}

Our findings contribute an empirical perspective to the relationship between grammaticalization and frequency, which has been discussed intensively (\shortciteNP<e.g.,>[cf.~p.~173]{DiMeola00}) but not been investigated in a rigorous way, as done here.

\bibliographystyle{apacite}
\bibliography{/Users/admin/Documents/workspace-python/backup/literature/Bibliography-general.bib}
\end{document}